# Invariant Gaussian Process Latent Variable Models and Application in Causal Discovery

**Kun Zhang**     **Bernhard Schölkopf**     **Dominik Janzing**

Max Planck Institute for Biological Cybernetics

Spemannstr. 38, 72076 Tübingen

Germany

## Abstract

In nonlinear latent variable models or dynamic models, if we consider the latent variables as confounders (common causes), the noise dependencies imply further relations between the observed variables. Such models are then closely related to causal discovery in the presence of nonlinear confounders, which is a challenging problem. However, generally in such models the observation noise is assumed to be independent across data dimensions, and consequently the noise dependencies are ignored. In this paper we focus on the Gaussian process latent variable model (GPLVM), from which we develop an extended model called invariant GPLVM (IGPLVM), which can adapt to arbitrary noise covariances. With the Gaussian process prior put on a particular transformation of the latent nonlinear functions, instead of the original ones, the algorithm for IGPLVM involves almost the same computational loads as that for the original GPLVM. Besides its potential application in causal discovery, IGPLVM has the advantage that its estimated latent nonlinear manifold is invariant to any nonsingular linear transformation of the data. Experimental results on both synthetic and real-world data show its encouraging performance in nonlinear manifold learning and causal discovery.

## 1 INTRODUCTION

Latent variable models, including dynamic models, aim to find a low-dimensional underlying manifold that helps understand the structure of the data. For instance, factor analysis and probabilistic principal component analysis (PPCA, Tipping & Bishop, 1999), as linear models, find the subspace of the common factors such that the error terms are uncorrelated from each other. Nonlinear latent variable models, such as the Gaussian process latent variable model (GPLVM, Lawrence, 2005), look for an intrinsically simple nonlinear manifold. In dynamic models, such as the Kalman filtering (Kalman, 1960) and Gaussian process dynamic model (GPDM, Wang et al., 2008), the latent processes are expected to be modelled well by a given temporal model.

In nonlinear latent variable models or dynamic models, usually the observation noise is assumed to be independent across data dimensions. This implicitly assumes that the interesting manifold coincides with the common subspace of the data, which is not necessarily true in practice. Consequently, the modelling performance and recovered latent variables are sensitive to the dependencies of the noise terms; in particular, they may be quite different for the data after linear transformations. This can be avoided by a nonlinear latent variable model that is able to adapt to arbitrary noise correlations, which will be developed in this paper.

On the other hand, the dependence between the noise terms is useful for data modelling, prediction, and causal discovery in the presence of latent common causes (confounders),[1] which is of particular interest to us. For example, in human motion tracking, different parts of humans may be closely related, or even causally related, almost instantaneously. By exploiting these relations, one could model or predict the joint distribution of the human parts better. From the causal discovery point of view (Spirtes et al., 2001; Pearl, 2000), making use of the error dependencies enables us to identify the linear causal relations among the observed variables, with the latent variables considered as confounders.

---

[1] Note that the noise dependence is generally very different from that of the observed variables. Generally speaking, due to the effect of the latent variables, the latter would be much more complex and less structured than the former. The latter can be directly estimated from the data and used to model the data structure; for example, in the motion tracking problem, the correlations between the right-side and the left-wide of the body were directly exploited in Xu and Li (2009), to reduce the computational load and improve the tracking accuracy.

In this paper the latent variable model of interest is GPLVM, a fully probabilistic one. Due to the Gaussian process (GP) prior (Rasmussen & Williams, 2006) on the nonlinear mapping, it is able to control the complexity of this mapping automatically, and can easily give probabilistic reconstructions of the data. It has been reported to work well with small data sets (Lawrence, 2005). However, to avoid the computational difficulties associated with a large kernel matrix which is caused by a dependent noise model, GPLVM assumes that the noise is independent across the dimensions. We circumvent these difficulties by putting the GP prior on a certain linear transformation of the latent functions. Consequently, we extend GPLVM (as well as GPDM, which is a combination of GPLVM with the GP prior on the temporal structure of the latent variables) to allow arbitrary noise covariances, while the algorithm still has similar computational loads as the original GPLVM.

The extension of GPLVM proposed by us has some appealing properties. First, its modelling performance and estimated latent variables are invariant to any nonsingular square linear transformation of the original data. Second, it admits a hierarchical way to explain the data generating process: the latent variables are a common cause to all observed variables, and the estimated noise dependencies imply further linear relations between the observed variables. If one believes such relations to be causal, the proposed method provides a suitable tool for causal discovery in the presence of nonlinear confounders.

## 2 NONLINEAR LATENT VARIABLE MODELS AND CAUSAL DISCOVERY

Let $\mathbf{y}_t = (y_{1t}, y_{2t}, ..., y_{Dt})^T$, $t = 1, 2, ..., N$, be a sequence of vector-valued observed variables. The nonlinear latent variable model considered here is

$$\mathbf{y}_t = \mathbf{g}(\mathbf{x}_t; \boldsymbol{\theta}) + \mathbf{e}_t, \quad (1)$$

where $\mathbf{x}_t = (x_{1t}, x_{2t}, ..., x_{dt})^T$ are latent/hidden variables, and the observation noise $\mathbf{e}_t$ is independent from the latent variables. In dynamic models, there is another model to specify the dynamic structure of the hidden variables. The $k$th order dynamic can be described as $\mathbf{x}_t = \mathbf{f}(\mathbf{x}_{t-1}, ..., \mathbf{x}_{t-k}; \boldsymbol{\eta}) + \boldsymbol{\varepsilon}_t$, where $\boldsymbol{\varepsilon}_t$ denotes the process noise.

### 2.1 RELATION TO CAUSAL DISCOVERY WITH NONLINEAR CONFOUNDERS

In some scenarios, we are more interested in the linear (causal) relations between the observed variables $y_{it}$ than the common effect of the latent variables. To this end, we describe the generating process of $\mathbf{y}_t$ as

$$\mathbf{y}_t = \mathbf{B}\mathbf{y}_t + \tilde{\mathbf{g}}(\mathbf{x}_t; \boldsymbol{\theta}) + \tilde{\mathbf{e}}_t, \quad (2)$$

where $\mathbf{B}$ contains the coefficients of the linear instantaneous influences, and the entries on its diagonal are constrained to be zero. If $\mathbf{B}$ can be transformed to strict low-triangularity by row and column permutations, its implied linear causal relations are acyclic. The model above is an extension of the linear causal structural equation model (Pearl, 2000) in that it has incorporated the effect of latent variables. Here it is assumed that the disturbances $\tilde{e}_{it}$ are independent from the latent variables (or confounders), as well as from each other.

Equation (2) can be written as

$$\begin{aligned}
(\mathbf{I} - \mathbf{B})\mathbf{y}_t &= \tilde{\mathbf{g}}(\mathbf{x}_t; \boldsymbol{\theta}) + \tilde{\mathbf{e}}_t \\
\Rightarrow \mathbf{y}_t &= (\mathbf{I} - \mathbf{B})^{-1} \cdot \tilde{\mathbf{g}}(\mathbf{x}_t; \boldsymbol{\theta}) + (\mathbf{I} - \mathbf{B})^{-1} \cdot \tilde{\mathbf{e}}_t \\
\Rightarrow \mathbf{y}_t &= \mathbf{g}(\mathbf{x}_t; \boldsymbol{\theta}) + \mathbf{e}_t,
\end{aligned}$$

where $\mathbf{g} \triangleq (\mathbf{I} - \mathbf{B})^{-1} \cdot \tilde{\mathbf{g}}$ and $\mathbf{e}_t \triangleq (\mathbf{I} - \mathbf{B})^{-1} \cdot \tilde{\mathbf{e}}_t$. This is exactly the latent variable model given by (1), but the noise terms $e_{it}$ are generally not independent.

One can then see that the linear causal relations between the observed variables are implied in the structure of the noise. Suppose that the noise in the latent variable model (1) has been estimated. If the disturbances $\tilde{e}_{it}$ are non-Gaussian, according to the independent component analysis (ICA, Hyvärinen et al., 2001) theory, $(\mathbf{I} - \mathbf{B})$ can be found up to certain permutation and scaling indeterminacies. Moreover, if the causal relations are acyclic, $\mathbf{B}$ is fully identifiable (Shimizu et al., 2006). In fact, this condition can be relaxed; under some assumptions, $\mathbf{B}$ may be identifiable even if there exist feedbacks between $y_{it}$ (Lacerda et al., 2008).

If $\tilde{e}_{it}$ are Gaussian, the observation noise $\mathbf{e}_t$ is Gaussian with the covariance matrix $\text{cov}(\mathbf{e}_t) = (\mathbf{I}-\mathbf{B})^{-1}\boldsymbol{\Sigma}_{\tilde{\mathbf{e}}}(\mathbf{I}-\mathbf{B})^{-T}$ (where $\boldsymbol{\Sigma}_{\tilde{\mathbf{e}}}$ is the covariance matrix of $\tilde{\mathbf{e}}_t$, which is diagonal). In this case, $(\mathbf{I} - \mathbf{B})$, as well as $\mathbf{B}$, is not identifiable, i.e., the whole causal model can not be fully determined. One can then use conditional independence-based approaches to find the Markov equivalence class of the causal model (Spirtes et al., 2001; Pearl, 2000), or the linear-Gaussian Markov network (Lauritzen, 1996) of the observed variables, with the effect of the latent variables excluded.

## 2.2 A GENERAL PROCEDURE FOR LINEAR CAUSAL DISCOVERY WITH NONLINEAR CONFOUNDERS

To estimate the linear (causal) relations between observed variables, with the latent variables considered as confounders, we first need to estimate the latent variable model (1) which is able to adapt to arbitrary noise dependencies. The estimated noise terms $\hat{e}_{it}$ are then obtained.

Next, depending on the problem and target, one can find the graphical model or causal model implied by the errors $\hat{e}_{it}$. One can always obtain the Gaussian Markov graph of $y_{it}$, which is implied by the precision matrix (the inverse covariance matrix) of $\hat{e}_{it}$. If we would like to find an acyclic causal model, we need to test if $\hat{e}_{it}$ are Gaussian. If they are, traditional conditional independence-based approaches can be used to find the Markov equivalence class of the causal model. If they are not, one can apply the linear, non-Gaussian, and acyclic model (LiNGAM, Shimizu et al., 2006) based analysis on $\hat{\mathbf{e}}_t$ to find the causal influence matrix $\mathbf{B}$. The case where the influences are not necessarily acyclic is more difficult to handle, although there exist certain methods, such as that proposed in Lacerda et al. (2008), to identify $\mathbf{B}$.

## 2.3 RELATION TO PREVIOUS WORK

Causal discovery in the presence of confounders is a very challenging problem, since it is difficult to find the confounders. In the linear confounder case, Hoyer et al. (2008) exploited overcomplete ICA to solve this problem. In the nonlinear case, Janzing et al. (2009) proposed a heuristic but useful method to examine if there exists a nonlinear confounder for two given observed variables. The model they exploited is essentially a nonlinear latent variable model with independent noise terms, and it inspired our work.

## 3 INVARIANT GAUSSIAN PROCESS LATENT VARIABLE MODELS

We aim to develop a reliable nonlinear latent variable model which can adapt to arbitrary noise correlations. Now the problem is to find a suitable form for the nonlinear mapping $\mathbf{g}$ in (1). Recently, the GP prior has been adopted to model it, resulting in GPLVM (Lawrence, 2004; Lawrence, 2005). This model is typically different from traditional nonlinear dimension reduction models in that it marginalizes the mapping instead of the latent variable. The latent variables are estimated together with the parameters appearing in the GP prior by maximizing the marginal likelihood.

In the original GPLVM (Lawrence, 2005), the GP prior is directly put on the nonlinear function $\mathbf{g}$ in (1). If the noise terms $e_{it}$ are dependent, the involved kernel matrix is of the dimensionality $DN \times DN$, which is very large and causes computational difficulties. Hence, the noise $\mathbf{e}_t$ in GPLVM is assumed to have uncorrelated Gassian components (or more frequently, to have isotropic Gaussian components). As explained in Section 2, this prevents GPLVM from being a suitable tool for causal discovery with nonlinear confounders. To tackle this problem, we extend GPLVM to allow arbitrary noise correlations. From the nonlinear dimension reduction point of view, the latent nonlinear manifold estimated by the extended model is invariant to any nonsingular square linear transformation of the data. Hence we name the extended model invariant GPLVM (IGPLVM). Two approaches for IGPLVM are given below.

### 3.1 Approach I: AN EFFICIENT SCHEME

#### 3.1.1 Model Formulation

Here we still assume that the noise $\mathbf{e}_t$ is Gaussian; treatment of the non-Gaussian noise will be discussed later. For a parsimonious parametrization, we use the Cholesky factorization to decompose the noise covariance matrix as $\text{cov}(\mathbf{e}_t) = \mathbf{L}\mathbf{L}^T$, where $\mathbf{L}$ is a lower-triangular matrix with strictly positive diagonal entries. That is, the noise is written as

$$\mathbf{e}_t = \mathbf{L}\mathbf{e}_t^*, \quad (3)$$

where $\mathbf{e}_t^*$ has independent components with an identity covariance matrix.

Combining (1) and (3) gives

$$\mathbf{y}_t = \mathbf{g}(\mathbf{x}_t; \boldsymbol{\theta}) + \mathbf{L}\mathbf{e}_t^*$$
$$\Rightarrow \quad \mathbf{y}_t = \mathbf{L}\mathbf{y}_t^I, \quad \text{where } \mathbf{y}_t^I \triangleq \mathbf{L}^{-1}\mathbf{g}(\mathbf{x}_t; \boldsymbol{\theta}) + \mathbf{e}_t^*. \quad (4)$$

Let $\mathbf{Y}_i$ be the column vector of the values of $y_{it}$, i.e., $\mathbf{Y}_i = (y_{i1}, ..., y_{iN})^T$, and $\mathbf{Y}$ be constructed by stacking all $\mathbf{Y}_i$ into a long vector, meaning $\mathbf{Y} = (\mathbf{Y}_1^T, ..., \mathbf{Y}_D^T)^T$. Similarly $\mathbf{Y}^I$ and $\mathbf{X}$ are constructed.

If one assumes that the components of $\mathbf{L}^{-1}\mathbf{g}(\mathbf{x}_t; \boldsymbol{\theta})$, as nonlinear functions of $\mathbf{x}_t$, can always be modelled with independent GP priors, i.e., we have $p(\mathbf{Y}^I|\mathbf{X}) = \Pi_{i=1}^D p(\mathbf{Y}_i^I|\mathbf{X})$, $\mathbf{y}_t^I$ can then be modelled by the original GPLVM (Lawrence, 2005), with independent noise terms and the noise variance fixed to one. Since $\mathbf{y}_t = \mathbf{L}\mathbf{y}_t^I$, we have $\mathbf{Y} = (\mathbf{L} \otimes \mathbf{I}_N)\mathbf{Y}^I$, where $\otimes$ denotes the Kronecker product and $\mathbf{I}_N$ denotes the $N$-dimensional identity matrix. Consequently, $\log p(\mathbf{Y}|\mathbf{X}) = \log p(\mathbf{Y}^I|\mathbf{X}) - N \log |\mathbf{L}|$. Finally, the

data log likelihood of (4) is

$$\begin{aligned}
l_1 &= \log p(\mathbf{Y}|\mathbf{X},\mathbf{L},r,\gamma) \\
&= \log p(\mathbf{Y}^I|\mathbf{X},r,\gamma) - N\log|\mathbf{L}| \\
&= -N\log|\mathbf{L}| - \frac{1}{2}\sum_{i=1}^{D}\mathbf{Y}_i^{IT}\mathbf{K}_{\mathbf{y}^I}^{-1}\mathbf{Y}_i^I \\
&\quad - \frac{D}{2}\log|\mathbf{K}_{\mathbf{y}^I}| - \frac{DN}{2}\log(2\pi),
\end{aligned} \quad (5)$$

where $r$ and $\gamma$ are parameters involved in the kernel matrix $\mathbf{K}_{\mathbf{y}^I}$, which is given below. Bear in mind that the noise in $y_{it}^I$ has a unit variance; we can then use a "RBF kernel" of the following form for $\mathbf{Y}_i^I$:[2]

$$k_{\mathbf{y}^I}(\mathbf{x}_t, \mathbf{x}_{t'}) = r \cdot \exp\left(-\frac{\gamma}{2}||\mathbf{x}_t - \mathbf{x}_{t'}||^2\right) + \delta_{\mathbf{x}_t, \mathbf{x}_{t'}}, \quad (6)$$

where $k_{\mathbf{y}^I}(\mathbf{x}_t, \mathbf{x}_{t'})$ is the $(t, t')$th entry of $\mathbf{K}_{\mathbf{y}^I}$. Parameters $\mathbf{X}$, $\mathbf{L}$, $r$, and $\gamma$ can then be estimated by maximizing the likelihood.

Thanks to the linear transformation $\mathbf{L}$ in (4), one can treat the components of $\mathbf{y}_t^I$ independently, and consequently the likelihood (5) involves the $N \times N$ kernel matrix $\mathbf{K}_{\mathbf{y}^I}$. If one directly applies independent GP priors on the components of $\mathbf{g}(\mathbf{x}_t; \boldsymbol{\theta})$ (as in the original GPLVM), the kernel matrix appearing in the likelihood would be of the dimensionality $DN \times DN$.

### 3.1.2 Learning and Discussions

Define $\widetilde{\mathbf{L}} \triangleq \mathbf{L}^{-1}$. It is also lower-triangular with positive diagonal entries. We alternate between maximizing the likelihood (5) w.r.t. $r$, $\gamma$, and $\mathbf{X}$ using the scaled conjugate gradient (SCG), which is similar to the parameter estimation procedure in the original GPLVM, and that w.r.t. $\widetilde{\mathbf{L}}$, which is obtained in closed-form in each iteration. Let $\mathbf{y}^I \triangleq (\mathbf{Y}_1^I, \mathbf{Y}_2^I, ..., \mathbf{Y}_D^I)^T$, which is of the size $D \times N$, and similarly for $\mathbf{y}$. The second term of $l_1$ given in (5) can then be written as $-\frac{1}{2}\text{Tr}(\widetilde{\mathbf{L}}\mathbf{y}\mathbf{K}_{\mathbf{y}^I}^{-1}\mathbf{y}^T\widetilde{\mathbf{L}}^T)$. In each update iteration, at the stationary point w.r.t. $\widetilde{\mathbf{L}}$, we have

$$\frac{\partial l_1}{\partial \widetilde{\mathbf{L}}} = N\widetilde{\mathbf{L}}^{-T} - \widetilde{\mathbf{L}}(\mathbf{y}\mathbf{K}_{\mathbf{y}^I}^{-1}\mathbf{y}^T) = 0.$$

Multiplying the above equation by $\widetilde{\mathbf{L}}^T$ from the right gives $\widetilde{\mathbf{L}}\mathbf{y}\mathbf{K}_{\mathbf{y}^I}^{-1}\mathbf{y}^T\widetilde{\mathbf{L}}^T = N\mathbf{I}_N$. That is, given the estimate of the remaining parameters, the optimal value of $\widetilde{\mathbf{L}}$ is the inverse of the (lower-triangular) Cholesky factorization matrix of $\frac{1}{N}\mathbf{y}\mathbf{K}_{\mathbf{y}^I}^{-1}\mathbf{y}^T$. In our implementation, PCA is used for initialing the latent variables $\mathbf{x}_t$, and both $r$ and $\gamma$ are initialized to one.

---

[2]Strictly speaking, $y_{it}^I, i = 1, 2, ..., D$, may have different signal-to-noise ratios (SNR's), and consequently the parameter $r$ in the kernel may vary for different dimensions of $y_{it}^I$. Here for simplicity we assume that all $y_{it}^I$ have the same SNR, like in Lawrence (2005); Wang et al. (2008).

Compared to the scaled GPLVM (Grochow et al., 2004), Approach *I* for IGPLVM involves $D(D+1) - D = D(D-1)/2$ more parameters, due to $\mathbf{L}$. But since these parameters are obtained in closed-form in each iteration, the computational load remains almost the same.

### 3.2 APPROACH *II*: AN EXACT BUT LESS EFFICIENT TREATMENT

#### 3.2.1 Model Formulation

In the original GPLVM (Lawrence, 2005), as well as GPDM (Wang et al., 2008), the components of $\mathbf{g}$ are treated independently; in Approach *I* for IGPLVM (Subsection 3.1), we put independent GP priors on the components of $\mathbf{L}^{-1}\mathbf{g}(\mathbf{x}_t; \boldsymbol{\theta})$ in (4). In practice these nonlinear functions are not necessarily independent.

To make the modelling performance of $\mathbf{g}$ completely insensitive to linear nonsingular transformations of the data, we further simply model $\mathbf{g} = (g_1, ..., g_D)^T$ as a linear transformation $\mathbf{R}$ of some functions with independent GP priors, denoted by $\mathbf{g}^* = (g_1^*, ..., g_D^*)$,[3] i.e.,

$$\mathbf{g}(\mathbf{x}_t) = \mathbf{R} \cdot \mathbf{g}^*(\mathbf{x}_t; \boldsymbol{\theta}). \quad (7)$$

This idea has been used in the semiparametric latent factor models for GP regression with multiple response variables (Teh & Seeger, 2005). However, the linear transformation $\mathbf{R}$ adopted here is different from Teh and Seeger (2005) in two aspects. First, for computational reasons, we constrain $\mathbf{R}$ to be square. Second, we give $\mathbf{R}$ a parsimonious parametrization. Without loss of generality, we parameterize $\mathbf{R}$ as a lower-triangular matrix with positive diagonal entries. This scheme has been frequently used to parameterize the loading matrix in factor analysis; for instance, see Geweke and Zhou (1996).

Combining (4) and (7) gives the observation equation:

$$\mathbf{y}_t = \mathbf{R} \cdot \mathbf{g}^*(\mathbf{x}_t; \boldsymbol{\theta}) + \mathbf{L}\mathbf{e}_t^*. \quad (8)$$

We construct $\mathbf{G}_i^*$ and $\mathbf{G}^*$ from $g_{it}^*$ in the way that $\mathbf{Y}$ and $\mathbf{X}$ were constructed in Subsection 3.1.1. Let $\mathbf{K}_{\mathbf{g}^*}$ be the kernel matrix shared by all $g_i^*, i = 1, ..., D$. We have $p(\mathbf{G}_i^*) \sim \mathcal{N}(\mathbf{0}, \mathbf{K}_{\mathbf{g}^*})$. As $g_i^*$ are assumed to have independent priors, one can see $p(\mathbf{G}^*) \sim \mathcal{N}(\mathbf{0}, \mathbf{K}_{\mathbf{G}^*})$, where $\mathbf{K}_{\mathbf{G}^*} = \mathbf{I}_D \otimes \mathbf{K}_{\mathbf{g}^*} = \text{diag}(\mathbf{K}_{\mathbf{g}^*}, ..., \mathbf{K}_{\mathbf{g}^*})$. We use a "RBF kernel" for the kernel shared by all $g_i^*$:

$$k_{\mathbf{g}^*}(\mathbf{x}_t, \mathbf{x}_{t'}) = \exp\left(-\frac{\gamma}{2}||\mathbf{x}_t - \mathbf{x}_{t'}||^2\right), \quad (9)$$

---

[3]Note that the diagonal scaling matrix in the scaled GPLVM (Grochow et al., 2004) or GPDM (Wang et al., 2008) can be considered a special case of such a linear transformation.

As in Lawrence (2005), there is a redundancy between $\gamma$ and the overall scale of $\mathbf{x}_t$; however, we still keep $\gamma$, because it is expected to compensate the overall scale of $\mathbf{x}_t$ quickly.

Let $\mathbf{y}_t^* \triangleq \mathbf{g}^*(\mathbf{x}_t, \boldsymbol{\theta}) + \mathbf{R}^{-1}\mathbf{L}\mathbf{e}_t^*$. Equation (8) can then be written as $\mathbf{y}_t = \mathbf{R}\mathbf{y}_t^*$. For simplicity of the notation, we define $\mathbf{L_R} \triangleq \mathbf{R}^{-1}\mathbf{L}$ and $\widetilde{\mathbf{R}} \triangleq \mathbf{R}^{-1}$. Let $\boldsymbol{\Sigma} = \mathbf{L_R}\mathbf{L_R}^T$. One can see[4]

$$p(\mathbf{Y}^*) \sim \mathcal{N}(\mathbf{0}, \mathbf{K_{Y^*}}), \text{ with } \mathbf{K_{Y^*}} = \mathbf{K_{G^*}} + \boldsymbol{\Sigma} \otimes \mathbf{I}_N.$$

Furthermore, since $\mathbf{y}_t = \mathbf{R}\mathbf{y}_t^*$, we have $\mathbf{Y} = (\mathbf{R} \otimes \mathbf{I}_N)\mathbf{Y}^*$, such that $\log p(\mathbf{Y}|\mathbf{X}) = \log p(\mathbf{Y}^*|\mathbf{X}) - N \log|\mathbf{R}|$. Therefore, the data log likelihood is

$$\begin{aligned} l_2 &= \log p(\mathbf{Y}|\mathbf{X}, \mathbf{L_R}, \widetilde{\mathbf{R}}, \gamma) \\ &= -\frac{1}{2}\mathbf{Y}^{*T}\mathbf{K_{Y^*}}^{-1}\mathbf{Y}^* - \frac{1}{2}\log|\mathbf{K_{Y^*}}| \\ &\quad + N\log|\widetilde{\mathbf{R}}| - \frac{DN}{2}\log 2\pi. \end{aligned} \quad (10)$$

### 3.2.2 Learning and discussions

All involved parameters, including $\mathbf{X}, \mathbf{L_R}, \widetilde{\mathbf{R}}$, and $\gamma$, are learned by maximizing the likelihood (10). We divide all parameters into two subsets. One contains free parameters in $\widetilde{\mathbf{R}}$ and $\mathbf{L_R}$ (both of which are constrained to be lower-triangular), and the other contains $\mathbf{X}$ and $\gamma$, which is involved in the kernel defined in (9). We alternatively update them by maximizing (10) using SCG for 10 iterations.

This approach is not suitable for relatively large datasets due to the high computational load in calculating the inverse and determinant of $\mathbf{K_{Y^*}}$, whose dimensionality is $DN \times DN$. Generally speaking, the complexity of finding $\mathbf{K_{Y^*}}^{-1}$ or $\det(\mathbf{K_{Y^*}})$ would be $\mathcal{O}(D^3 N^3)$. One can exploit the informative vector machine (IVM, Lawrence et al., 2003) framework or other more recent techniques to select an active set of the training data and reduce the computational load, by extending the methods given by Lawrence (2005); Teh and Seeger (2005); Lawrence (2007). Furthermore, Approach *II* has $D(D+1)/2 - 1$ more parameters than Approach *I* given in Subsection 3.1. Consequently, it might be more prone to local optima and overfitting. However, this approach is still presented here, as an exact formulation for IGPLVM. In practice we found that Approach *I* plays a nice trade-off of the model capacity and simplicity.

---

[4] We use the kernel matrix of $\mathbf{Y}^*$ because it is sparse and has a special structure, which may make it easier to calculate its inverse and determinant.

## 3.3 NON-GAUSSIAN CASE

As claimed in Section 2, in the case that the observation noise is non-Gaussian, it might be possible to fully identify the linear causal relations among the observed variables, with the latent variables treated as confounders. Here we give the basic idea of how to estimate IGPLVM with non-Gaussian noise.

When the noise is Gaussian, the marginal distribution $p(\mathbf{Y}|\mathbf{X}, \boldsymbol{\theta})$ is in closed form; see (5) or (10). However, generally speaking, with a non-Gaussian distribution for $e_{it}$, this marginal distribution could not be expressed in closed-form any more, causing difficulties in parameter estimation and inference. A straightforward way is to resort to the Laplace approximation, which approximates the posterior of the latent functions with a Gaussian density. Although theoretically rather simple, this approach would lead to complicated algorithms for IGPLVM.

### 3.3.1 Two-Step Method

For practical reasons we propose a simple heuristic two-step method to solve this problem. In the first step, we decompose the data into the nonlinear manifold and the noise effect, i.e., estimate the latent processes $\mathbf{x}_t$ and the noise $\mathbf{e}_t$, with the Gaussianity-based IGPLVM. In the second step we consider the non-Gaussianity of the noise. We test if $\hat{e}_{it}$ are Gaussian. If they are not, apply ICA (Hyvärinen et al., 2001), which makes use of the non-Gaussianity, to further find their linear structure, i.e., to estimate the matrix $\mathbf{L}$ in (3).[5] (In this case, $\mathbf{L}$ is a full matrix, instead of a lower-triangular one.)

## 4 EXPERIMENTS

### 4.1 SIMULATIONS

Invariant GPDM (IGPDM) can be obtained straightforwardly by combining the proposed IGPLVM with the autoregressive GP regression model for the latent processes. The temporal structure of those processes helps estimate them more reliably. We use simulations to compare the performance of IGPDM with the original GPDM (Wang et al., 2008) and demonstrate the behavior of the two-step method for IGPDM with non-Gaussian noise proposed in Subsection 3.3.

The simulated data were constructed in the following way. The two-dimensional underlying latent pro-cess $\mathbf{x}_t$ smoothly moved on a circle; see Figure 1(a) for its trajectory. The sample size was $N = 400$. The eight-dimensional observed process $\mathbf{y}_t$ was generated according to the nonlinear latent variable model (1); components of $\mathbf{g}$ were random mixtures of various smooth functions, including linear, quadratic, third-order polynomial, and tanh functions. The error $\mathbf{e}_t$ has a special structure; it was constructed as a linear transformation of some independent variables, i.e., $\mathbf{e}_t = \mathbf{A}\mathbf{s}_t$. $s_{it}$ were obtained by passing independent and i.i.d. Gaussian samples through power nonlinearities with the exponent between 1.5 and 2. Thus the noise is non-Gaussian (supper-Gaussian, in fact). The variances of $s_{it}$ were randomly chosen between 0.2 and 1. The mixing matrix $\mathbf{A}$ was lower-triangular with random entries (between -1 and 1) in the strictly lower-triangular part, and the entries on its diagonal were set to one. Consequently, if $\mathbf{x}_t$ is considered as a confounder for all $y_{it}$, the causal relations between $y_{it}$ are linear, acyclic, and fully connected. Furthermore, to make some components of $\mathbf{e}_t$ significantly dependent on each other, we made three rows of $\mathbf{A}$ have similar but different entries.

---

[5] This method is not efficient in the statistical sense. However, it is expected to recover the nonlinear manifold and the noise approximately if the noise is not extremely non-Gaussian. We will use simulations to illustrate this in Subsection 4.1.

Our goals were 1) to see if IGPDM could successfully estimate the nonlinear manifold and noise, 2) to examine if the two-step method for IGPDM with non-Gaussian noise could recover the noise structure, and 3) furthermore, to see if the causal relations between $y_{it}$ introduced by the noise could be identified by the causal discovery procedure given in Subsection 2.2. We used two measures for performance evaluation. One is the mean square error (MSE) between the true nonlinear functions $g_i$ and the estimated ones; it measures how accurately the nonlinear manifold is recovered. The other is the Amari performance index (Hyvärinen et al., 2001), denoted by $P_{err}$, which is frequently used for source separation performance evaluation. It is used here to measure how accurately the matrix $\mathbf{A}$, which generated the noise, can be estimated by applying ICA on $\hat{e}_{it}$ (which is step 2 of the two-step method given in Subsection 3.3). The smaller this index, the better the performance.

We repeated the simulations for 5 random replications (we did not run more replications because Approach II for IGPDM given in Subsection 3.2 is time-consuming). All methods were initialized by PCA. Table 1 reports the performance. One can see that due to the impact of the noise dependencies, it is not easy for the original GPDM to recover the smooth latent manifold. In contrast, incorporation of the noise structure enables the proposed methods, especially Approach I (Subsection 3.1), to estimate the latent variables as well as the noise very well. Approach II seems not as stable as Approach I, meaning that it may be prone to local optima or over-fitting. Approach I plays a nice trade-off of the simplicity and capacity of the model. The LiNGAM (Shimizu et al., 2006) software[6] was then used to analyze the estimated noise to find the causal relations between $y_{it}$. For the noise estimated by Approach I for IGPDM, in all five replications, the software always reported that $\hat{\mathbf{e}}_t$ nicely follows the LiNGAM model. In average 71% of the causal relations were recovered, and the chance of producing spurious relations was 7%. This is consistent with the good performance in the recovery of $\mathbf{A}$, as shown by the small values of $P_{err}$. For the noise estimated by the original GPDM, in all replications, LiNGAM reported that $\hat{\mathbf{e}}_t$ does not follow LiNGAM or that the causal relations are only somewhat acyclic.

Figure 1 (b,c,d) show the the learned latent processes in the first replication. Note that the comparison is fair: different methods employed the same prior on the latent process. Both methods for IGPDM successfully recovered the subspace of the latent process. Approach I was the best. We then applied a random rotation to the observed processes $\mathbf{y}_t$ and learned the models for another time. The leaned latent processes are shown by crosses. Clearly the results of IGPDM remained the same (note that for Approach II, the results slightly changed, which was caused by different parameter initializations.) However, the results learned by the original GPDM changed a lot. This verifies that the performance of the proposed IGPDM (as a special case of IGPLVM) is rotation-invariant.

### 4.2 ON MOTION TRACKING

We then applied IGPLVM on a human motion capture data set extracted from CMU mocap database. We used the data of two walk cycles (subject 7, trial 1) for analysis. The same data have been used by Wang et al. (2008). Our goal was to see if the proposed IGPDM could find appealing relations between different joints. To make such relations more obvious, we further simplified the skeleton used by Wang et al. (2008), only keeping the limbs, as shown in Figure 3. Finally 22 variables were used.

To save space, the learned latent process is not given. As it is difficult to believe that the joints have acyclic causal relations, we are interested in the learned noise covariances and the implied Gaussian Markov network. Figure 2 shows the precision matrices (inverse covariance matrices) of the observed variables, of the error estimated by Approach I for IGPDM, and of the error estimated by the original GPDM. Zero entries in the precision matrix indicate missing edges in the Markov network. One can see that the network implied by the observed data is very dense and complicated. In contrast, that implied by the error estimated by IG-PDM gives the most sparse and simple network. Figure 3 compares the corresponding Markov networks, in each of which 20 most significant connections are shown. As the subject in the dataset walked straight, the network produced by IGPDM (panel b), which is obviously more symmetric, seems more natural.

---

[6]http://www.cs.helsinki.fi/group/neuroinf/lingam/

Table 1: Model comparison with simulated data.

| Methods | Criteria | Run 1 | Run 2 | Run 3 | Run 4 | Run 5 | Average |
|---|---|---|---|---|---|---|---|
| GPDM | MSE of $\hat{\mathbf{g}}_t$ | 0.22 | 0.2 | 0.18 | 0.24 | 0.20 | $0.21 \pm 0.02$ |
| | $P_{err}$ of $\hat{\mathbf{e}}_t$ | 0.46 | 0.51 | 0.42 | 0.35 | 0.41 | $0.43 \pm 0.06$ |
| IGPDM (Approach *I*) | MSE of $\hat{\mathbf{g}}_t$ | 0.017 | 0.017 | 0.015 | 0.068 | 0.025 | $0.028 \pm 0.022$ |
| | $P_{err}$ of $\hat{\mathbf{e}}_t$ | 0.11 | 0.12 | 0.14 | 0.19 | 0.10 | $0.13 \pm 0.04$ |
| IGPDM (Approach *II*) | MSE of $\hat{\mathbf{g}}_t$ | 0.02 | 0.06 | 0.012 | 0.062 | 0.10 | $0.05 \pm 0.04$ |
| | $P_{err}$ of $\hat{\mathbf{e}}_t$ | 0.13 | 0.32 | 0.11 | 0.29 | 0.34 | $0.24 \pm 0.11$ |

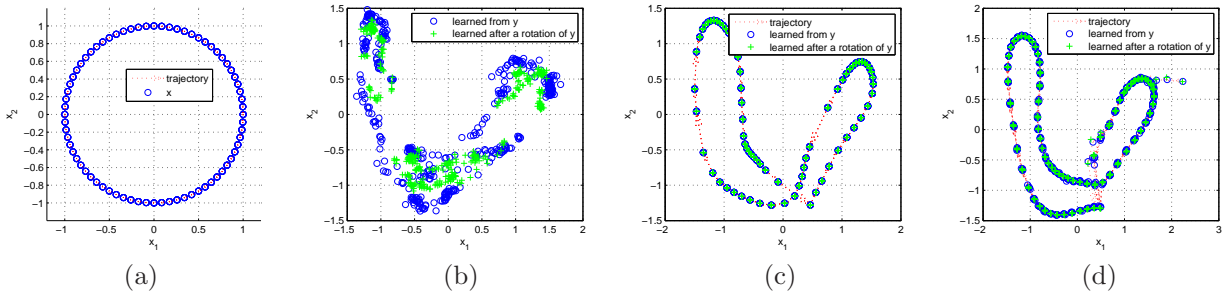

Figure 1: Scatter plot of the latent process $\mathbf{x}_t$. (a) That used to generate $\mathbf{y}_t$. (b) That estimated by original GPDM. (c) That estimated by Approach *I* for IGPDM. (d) That estimated by Approach *II* for IGPDM.

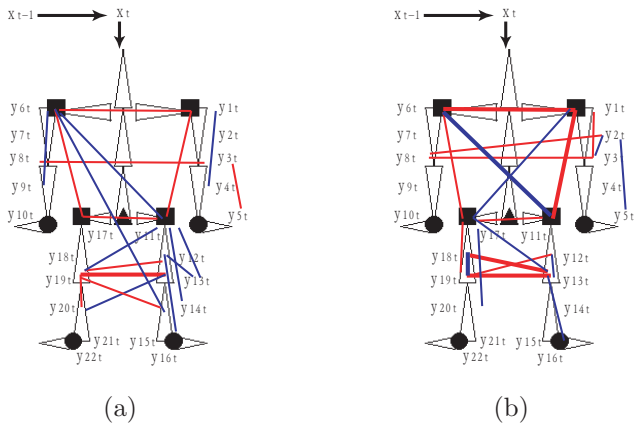

Figure 3: Estimated linear-Gaussian Markov networks of human joints during walking. Black lines, squares, and circles show the skeleton. Blue (red) lines indicate positive (negative) connections. The thicker the line, the stronger the connection. (a) That implied by the error estimated by GPDM. (b) That implied by the error estimated by IGPDM (Approach *I*).

## 5 DISCUSSIONS

In this paper we proposed an extension of the Gaussian process latent variable model (GPLVM), namely, invariant GPLVM (IGPLVM). It has the appealing property that its estimated nonlinear manifold is invariant to any nonsingular linear transformation of the data. Since this model allows arbitrary noise correlations, we show that it also provides a suitable tool to solve an inherently difficult causal discovery problem – causal discovery in the presence of nonlinear confounders. The proposed Approach *I* for IGPLVM is recommended; it involves almost the same complexity as the original GPLVM, and we found that in practice this approach, in combination with suitable priors on the temporal structure of the latent variables, is able to estimate the latent functions and noise structure (including the implied causal relations).

In the original GPLVM as well as the proposed extensions, the model parameters in the nonlinear mapping and the latent variables are estimated simultaneously. As the latent variables consist of many values, the learning process may be prone to multiple local optima and overfitting, if ones does not have suitable temporal constraints on the latent variables. As reported by Lawrence (2005), the result of GPLVM may be sensitive to initializations. This is inherently related to the difficulties in non-Gaussianity-based causal discovery with linear confounders (Hoyer et al., 2008). However, suitable priors on the temporal structure of the latent processes help alleviate this problem significantly, as shown in our experiments, in which au-toregressive Gaussian process models were used for the latent processes. To improve the applicability of IG-PLVM for causal discovery in various practical problems, one line of our future research is to investigate the possibility of developing more reliable algorithms for IGPLVM without temporal constraints on the latent variables.

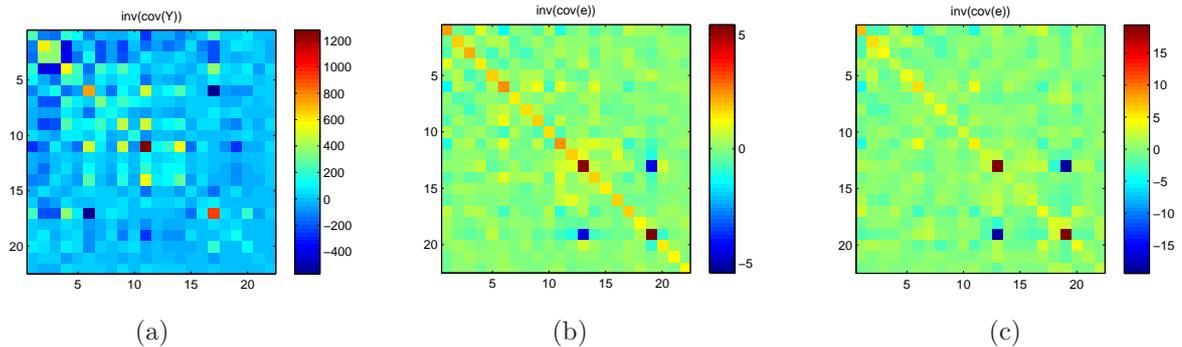

Figure 2: The precision matrix obtained by different methods. (a) That of the raw observed data. (b) That of the errors estimated by original GPDM. (c) That of the errors estimated by Approach *I* for IGPDM. Here each error term was normalized to unit variance.

## Acknowledgements

We would like to thank Aapo Hyvärinen and Jonas Peters for helpful discussions.

## References

Geweke, J., & Zhou, G. (1996). Measuring the pricing error of the arbitrage pricing theory. *Rev. Finan. Stud.*, *9*, 557–587.

Grochow, K., Martin, S., Hertzmann, A., & Popovic, Z. (2004). Style-based inverse kinematics. *ACM Transactions on Graphics*, *23*, 522–531.

Hoyer, P., Shimizu, S., Kerminen, A., & Palviainen, M. (2008). Estimation of causal effects using linear non-gaussian causal models with hidden variables. *International Journal of Approximate Reasoning*, *49*, 362–378.

Hyvärinen, A., Karhunen, J., & Oja, E. (2001). *Independent component analysis*. John Wiley & Sons, Inc.

Janzing, D., Peters, J., Mooij, J., & Schölkopf, B. (2009). Identifying confounders using additive noise models. *Proceedings of the 25th Conference on Uncertainty in Artificial Intelligence (UAI 2009)* (pp. 249–257).

Kalman, R. (1960). A new approach to linear filtering and prediction problems. *Journal of Basic Engineering*, *82*, 35–45.

Lacerda, G., Spirtes, P., Ramsey, J., & Hoyer, P. O. (2008). Discovering cyclic causal models by independent components analysis. *Proceedings of the 24th Conference on Uncertainty in Artificial Intelligence (UAI2008)*. Helsinki, Finland.

Lauritzen, S. (1996). *Graphical models*. Oxford.

Lawrence, N. (2004). Gaussian process latent variable models for visualization of high dimensional data. *Advances in Neural Information Processing Systems 16* (pp. 329–336). The MIT Press.

Lawrence, N. (2005). Probabilistic non-linear principal component analysis with Gaussian process latent variable models. *Journal of Machine Learning Research*, *6*, 1783–1816.

Lawrence, N. (2007). Learning for larger datasets with the Gaussian process latent variable model. *AISTATS 2007* (pp. 21–24). San Juan, Puerto Rico.

Lawrence, N., Seeger, M., & Herbrich, R. (2003). Fast sparse Gaussian process methods: The informative vector machine. *Advances in Neural Information Processing Systems 15* (pp. 609–616). The MIT Press.

Pearl, J. (2000). *Causality: Models, reasoning, and inference*. Cambridge: Cambridge University Press.

Rasmussen, C., & Williams, C. (2006). *Gaussian processes for machine learning*. Cambridge, Massachusetts, USA: MIT Press.

Shimizu, S., Hoyer, P., Hyvärinen, A., & Kerminen, A. (2006). A linear non-Gaussian acyclic model for causal discovery. *Journal of Machine Learning Research*, *7*, 2003–2030.

Spirtes, P., Glymour, C., & Scheines, R. (2001). *Causation, prediction, and search*. Cambridge, MA: MIT Press. 2nd edition.

Teh, Y. W., & Seeger, M. (2005). Semiparametric latent factor models. *Workshop on Artificial Intelligence and Statistics 10*.

Tipping, M., & Bishop, C. (1999). Probabilistic principal component analysis. *Journal of the Royal Statistical Society, Series B*, *61*, 611–622.

Wang, J., Fleet, J., & Hertzmann, A. (2008). Gaussian process dynamical models for human motion. *IEEE Transactions on Pattern Recognition and Machine Intelligence*, *30*, 283–298.

Xu, X., & Li, B. (2009). Exploit motion correlations for 3d articulated human tracking. *IEEE Transactions on Image Processing*, *16*, 838–849.